\pdfoutput=1

\documentclass[11pt]{article}

\usepackage[final]{acl}

\usepackage{times}
\usepackage{latexsym}
\usepackage[T1]{fontenc}
\usepackage[utf8]{inputenc}
\usepackage{microtype}
\usepackage{inconsolata}
\usepackage{graphicx}
\usepackage{booktabs}
\usepackage{tikz}
\usepackage{orcidlink}

\usetikzlibrary{positioning,arrows.meta,shapes,shapes.geometric,patterns,calc}
\usepackage{pgfplots}
\pgfplotsset{compat=1.17}
\usepackage{amsmath}
\usepackage{amsfonts}
\usepackage{algorithm, algorithmic}
\usepackage{listings}
\usepackage{float}

\newcommand{\sis}{\textsc{SIS}}
\newcommand{\TopK}{\operatorname{TopK}}
\newcommand{\Expand}{\operatorname{Expand}}
\newcommand{\LCA}{\operatorname{LCA}}
\newcommand{\softmax}{\operatorname{softmax}}
\newcommand{\sigmoid}{\operatorname{sigmoid}}

\title{NOEM\textsuperscript{3}A: a Neuro-symbolic Ontology-Enhanced Method \\
for Multi-intent understanding in Mobile Agents}

\author{
\textbf{Ioannis Tzachristas}\,\textsuperscript{1,2}\,\orcidlink{0009-0007-0523-2889},
\textbf{Aifen Sui}\,\textsuperscript{1}\thanks{Correspondence: \mbox{\texttt{aifen.sui@huawei.com}}.}\,\orcidlink{0009-0006-2094-9178} \\[2pt]
\textsuperscript{1}Huawei Heisenberg Research Center, Munich, Germany \\
\textsuperscript{2}Technical University of Munich, Germany 
}

\begin{document}
\maketitle

\begin{abstract}
Mobile agents must map natural-language requests to executable intents under tight latency and privacy constraints. Scaling the language model is often an inefficient way to improve this component. We present NOEM$^{3}$A, a lightweight neuro-symbolic layer that augments compact language models with an intent ontology. For each query, NOEM$^{3}$A retrieves a small ontology neighborhood, injects candidate action labels into the prompt and applies a token-level decoding prior toward valid labels. This injects symbolic intent structure into both input and output representations while keeping inference suitable for local interaction. We also use Semantic Intent Similarity (SIS), a hierarchy-aware diagnostic based on ontology depth, to capture semantic proximity when predicted intents differ lexically. Experiments on MultiWOZ 2.3 dialogues show that ontology augmentation consistently improves TinyLlama and Llama-3.2-3B on exact match and Slot-F1. Our results suggest that symbolic alignment is an effective strategy for accurate, responsive and privacy-preserving on-device NLU.
\end{abstract}
\section{Introduction}
Mobile assistants and GUI-based agents increasingly perform actions rather than only answer questions~\citep{tzachristas2024creating}: booking travel, creating reminders, ordering food, searching products or coordinating several tools. A single utterance may contain multiple intents~\citep{Gangadharaiah2019}, as in \emph{"find a hotel near the station, book a taxi and remind me before departure".} The language component must therefore produce a valid set of tool-facing labels, and it must do so fast enough for local interaction.

Large Language Models are effective semantic parsers, but they are not always the right deployment primitive for mobile agents. Cloud inference introduces privacy exposure and tail latency, while larger local models increase memory pressure and token-by-token compute. We study a complementary efficiency strategy: keep the neural model compact and supply an explicit symbolic action space at inference time.

We propose NOEM$^{3}$A{} (Neuro-symbolic Ontology-Enhanced Method for Multi-intent understanding in Mobile Agents). Following ontology/knowledge-guided NLP work~\citep{Pinhanez2021,He2021,Yu2022,He2023}, the method uses an ontology whose nodes are mobile-agent domains, categories and executable intents. At runtime, the system retrieves a query-specific subgraph and uses it as both input context and output constraint. This shifts part of the reasoning burden from parametric generation to a small, inspectable symbolic lookup. Related work further includes out-of-scope and zero-shot intent detection~\citep{Cavalin2020,Siddique2021}, LLM-assisted ontology construction and graph prompting~\citep{Vassilakis2025Ontology,Wen2024MindMap,Liu2025ORT}, knowledge-injected language models~\citep{Zhang2019} and dialogue-state methods used as MultiWOZ 2.3 context~\citep{hu2022icdst,lee2024seri,heck2023chatgpt,chung2023instructtods,li2024fnctod}.

This paper focuses on efficient inference. We use exact match and Slot-F1 as primary task measures. We also report Semantic Intent Similarity (\sis) as an auxiliary diagnostic for whether an error stays near the correct ontology branch, but we avoid treating it as a standalone claim of success.

Our contributions are: (i) a formalized ontology-guided inference layer for multi-intent mobile NLU; (ii) a latency-focused accounting that separates retrieval, decoding bias and neural inference; and (iii) an evaluation showing that symbolic guidance improves compact models without requiring a larger generator.

\begin{figure*}[t]
\centering
\includegraphics[width=0.83\linewidth]{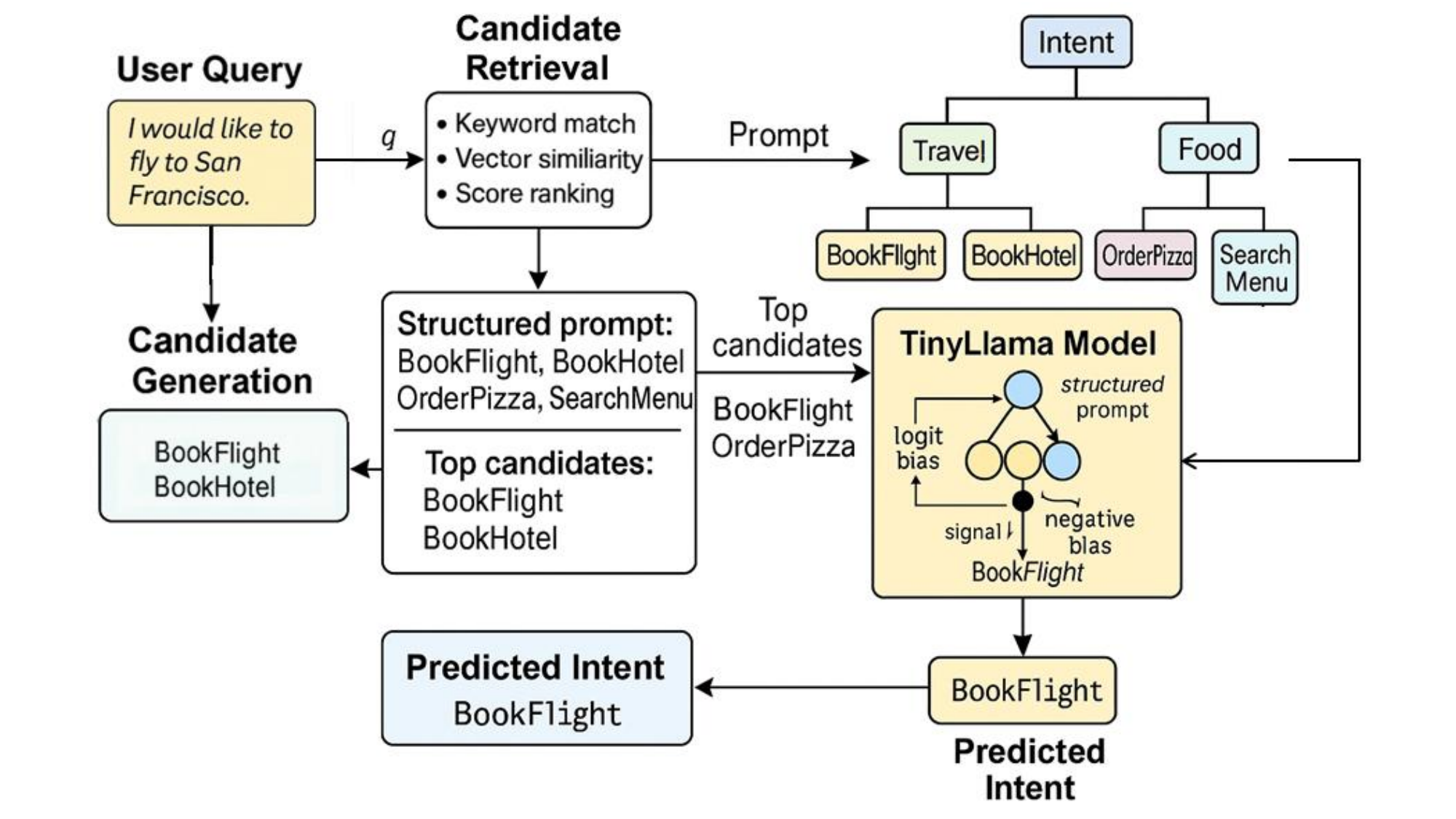}
\caption{The NOEM$^{3}$A{} pipeline. A user query is matched to candidate ontology nodes; the retrieved candidates are serialized into a structured prompt and used as a decoding prior so that the compact model emits ontology-valid intent labels.}
\label{fig:architecture}
\end{figure*}

\begin{figure*}[t]
    \centering
    \resizebox{0.98\linewidth}{!}{%
    \begin{tikzpicture}[
      every node/.style={font=\footnotesize, align=center},
      level distance=1.15cm,
      level 1/.style={sibling distance=40mm},
      level 2/.style={sibling distance=17mm},
      edge from parent/.style={draw,-Latex}
    ]
    \node[draw, rectangle, fill=gray!20] {All Intents}
        child { node[draw, rectangle, fill=blue!15] {Ticket\\Booking}
            child { node[draw, rectangle, fill=blue!5] {Flight\\Booking} }
            child { node[draw, rectangle, fill=blue!5] {Event\\Ticket} }
        }
        child { node[draw, rectangle, fill=orange!15] {Food\\Delivery}
            child { node[draw, rectangle, fill=orange!5] {Restaurant\\Order} }
            child { node[draw, rectangle, fill=orange!5] {Grocery\\Order} }
        }
        child { node[draw, rectangle, fill=green!15] {Holiday\\Planning}
            child { node[draw, rectangle, fill=green!5] {Itinerary\\Planning} }
            child { node[draw, rectangle, fill=green!5] {Hotel\\Search} }
        }
        child { node[draw, rectangle, fill=purple!15] {Online\\Shopping}
            child { node[draw, rectangle, fill=purple!5] {Product\\Search} }
            child { node[draw, rectangle, fill=purple!5] {Purchase\\Order} }
            child { node[draw, rectangle, fill=purple!5] {Order\\Tracking} }
        };
    \end{tikzpicture}}
    \caption{Excerpt from the ontology graph of mobile intents. Four top-level domains (Ticket Booking, Food Delivery, Holiday Planning, Online Shopping) are shown with some of their sub-intents. Each intent label is split over two lines for improved visual clarity.}
    \label{fig:ontology}
\end{figure*}

\section{Formal Method}
\paragraph{Intent ontology.}
We represent the action space as a directed ontology $\mathcal{O}=(V,E)$, where each node $v\in V$ is a domain, category or executable intent and edges encode typed relations such as \textsc{is-a} or \textsc{related-to}. Each node stores a label $\ell(v)$, description $d(v)$ and embedding
\begin{equation}
    e_v = f\bigl(\ell(v) \oplus d(v)\bigr)\in\mathbb{R}^{m},
\end{equation}
where $f$ is the same lightweight encoder used at inference time. The ontology therefore acts as both an action schema and a searchable memory over valid labels.

\paragraph{Subgraph retrieval.}
Given a query $q$, we compute $e_q=f(q)$ and rank ontology nodes by cosine similarity,
\begin{equation}
    s(q,v)=\frac{e_q^\top e_v}{\lVert e_q\rVert_2\lVert e_v\rVert_2}.
\end{equation}
The raw candidate set is $C_q=\TopK_{v\in V} s(q,v)$. We then retrieve a local neighborhood
\begin{equation}
    G_q=\Expand(\mathcal{O}, C_q; r),
\end{equation}
where $r$ controls optional parent, child or sibling expansion. In practice, $G_q$ is small and fixed, so the downstream prompt remains short even when the ontology grows.

\paragraph{Ontology-augmented prompt.}
The retrieved labels are serialized into a compact instruction. Abstractly,
\begin{equation}
    P_q = \bigl[q;\ell(v_1),\ldots,\ell(v_k)\bigr],
\end{equation}
where the concrete prompt asks the model to choose one or more intents from the listed labels. This changes the task from unconstrained generation to selection among plausible ontology labels. Unlike document retrieval methods~\citep{Lewis2020,Han2024,Edge2024}, the retrieved objects are not evidence passages but executable action types. Figure~\ref{fig:alignment} illustrates this token-to-node alignment.

\begin{figure}[t]
    \centering
    \resizebox{0.98\linewidth}{!}{%
    \begin{tikzpicture}[node distance=1mm and 2mm, every node/.style={font=\footnotesize}]
      \node[draw, rectangle, fill=gray!10] (token1) {Order};
      \node[draw, rectangle, fill=gray!10, right=of token1] (token2) {a};
      \node[draw, rectangle, fill=gray!10, right=of token2] (token3) {pizza};
      \node[draw, rectangle, fill=gray!10, right=of token3] (token4) {and};
      \node[draw, rectangle, fill=gray!10, right=of token4] (token5) {track};
      \node[draw, rectangle, fill=gray!10, right=of token5] (token6) {my};
      \node[draw, rectangle, fill=gray!10, right=of token6] (token7) {order};
      \node[below=1em of token4, align=center] {User Query \\Tokens};
      \node[draw, rectangle, fill=orange!10, below left=1.8cm and -0.2cm of token3] (nodeA) {\textit{Restaurant Order}};
      \node[draw, rectangle, fill=purple!10, below right=1.8cm and -0.2cm of token4] (nodeB) {\textit{Order Tracking}};
      \node[below=0.1cm of nodeA] {Ontology Nodes};
      \draw[-stealth, thick] (token3) -- (nodeA) node[midway, above, sloped] {\scriptsize 0.88};
      \draw[-stealth, thick] (token1) -- (nodeA) node[midway, above, sloped] {\scriptsize 0.75};
      \draw[-stealth, thick] (token7) -- (nodeB) node[midway, above, sloped] {\scriptsize 0.80};
      \draw[-stealth, thick] (token5) -- (nodeB) node[midway, above, sloped] {\scriptsize 0.79};
      \draw[dashed, -stealth, gray] (token7.south) to[out=-90, in=30] (nodeA.east);
      \draw[dashed, -stealth, gray] (token1.south) to[out=-90, in=150] (nodeB.west);
    \end{tikzpicture}}
    \caption{Semantic alignment between query tokens and ontology node embeddings. For the query "Order a pizza and track my order", food-ordering tokens align with \textit{Restaurant Order}, while the tracking context aligns with \textit{Order Tracking}. This alignment allows the retriever to select both intent nodes as relevant.}
    \label{fig:alignment}
\end{figure}

\paragraph{Token-level decoding prior.}
Let $T(G_q)$ be the set of tokenizer IDs appearing in labels from $G_q$. During decoding, we add a small bias to candidate-label tokens and a mild penalty to unrelated label tokens:
\begin{equation}
    b_t=\begin{cases}
        \beta, & t\in T(G_q),\\
        -\gamma, & t\in T(V\setminus G_q),\\
        0, & \text{otherwise,}
    \end{cases}
    \qquad
    z'_t=z_t+b_t .
\end{equation}
The next-token distribution is then $p_t=\softmax(z')_t$. We tune $\beta$ and $\gamma$ conservatively; the goal is to regularize labels rather than force an intent when the query does not support it. Figure~\ref{fig:logit_bias} shows the intended effect of the decoding prior.

\begin{figure}[t]
\centering
\resizebox{0.98\linewidth}{!}{%
\begin{tikzpicture}
\begin{axis}[
    ybar,
    ymin=0,
    ymax=1,
    width=0.95\linewidth,
    height=4.7cm,
    bar width=7pt,
    enlarge x limits=0.2,
    symbolic x coords={BookFlight, BookHotel, OrderPizza, CheckWeather, Other},
    xtick=data,
    xticklabel style={rotate=45, anchor=east, yshift=-0.2cm},
    ylabel={Logit Value},
    ymajorgrids=true,
    grid style={dashed,gray!30},
    legend style={at={(0.5,1.05)}, anchor=south, legend columns=2},
    legend cell align={left}
]
\addplot+[style={fill=gray!30}, postaction={pattern=north east lines}] coordinates {
    (BookFlight, 0.45)
    (BookHotel, 0.40)
    (OrderPizza, 0.50)
    (CheckWeather, 0.40)
    (Other, 0.65)
};
\addplot+[style={fill=blue!50}] coordinates {
    (BookFlight, 0.75)
    (BookHotel, 0.70)
    (OrderPizza, 0.30)
    (CheckWeather, 0.20)
    (Other, 0.45)
};
\legend{Original logits, After logit bias}
\end{axis}
\end{tikzpicture}}
\caption{Effect of logit biasing in ontology-augmented inference. Ontology-relevant intents like \texttt{BookFlight} and \texttt{BookHotel} receive higher logits, while less relevant or unrelated intents are suppressed.}
\label{fig:logit_bias}
\end{figure}

\paragraph{Auxiliary classifier.}
When labeled data are available, we attach a multi-label classifier to the pooled hidden state $h_q$:
\begin{equation}
    p^{\mathrm{clf}}=\sigmoid(W h_q+a),
\end{equation}
\begin{equation}
\begin{split}
    \mathcal{L}_{\mathrm{clf}}=-\sum_{v\in V}\bigl[&y_v\log p_v \\
    &+(1-y_v)\log(1-p_v)\bigr].
\end{split}
\end{equation}
Generated labels and classifier labels are merged as
\begin{equation}
    \hat{Y}=Y_{\mathrm{gen}} \cup \{v\in V: p^{\mathrm{clf}}_v>\tau\}.
\end{equation}
This optional head is useful when a mobile platform has curated logs or simulator-generated labels.

\paragraph{Hierarchy-aware diagnostic.}
For analysis only, we compute a taxonomy score inspired by Wu and Palmer similarity~\citep{wu1994verbs}. For predicted node $u$ and gold node $v$,
\begin{equation}
    \sis(u,v)=\frac{2\,\mathrm{depth}(\LCA(u,v))}{\mathrm{depth}(u)+\mathrm{depth}(v)}.
\end{equation}
For multi-intent sets, we compute a weighted bipartite matching between predicted and gold nodes and average matched similarities. Figure~\ref{fig:sis-final} gives an example calculation. Because \sis{} depends on the chosen ontology topology, it is a debugging signal rather than a cross-system leaderboard metric.

\begin{figure*}[t]
\centering
\resizebox{0.9\linewidth}{!}{%
\begin{tikzpicture}[
  level distance=18mm,
  sibling distance=40mm,
  every node/.style = {
    draw, rounded corners, minimum width=2.0cm, minimum height=9mm,
    font=\footnotesize, align=center
  },
  edge from parent/.style = {draw, -Latex, thick},
  edge from parent path={(\tikzparentnode.south) -- (\tikzchildnode.north)},
  level 1/.style = {sibling distance=70mm},
  level 2/.style = {sibling distance=45mm},
  level 3/.style = {sibling distance=23mm}
]
\begin{scope}[xshift=-1.5cm]
\node (root) {Intent \\ (depth = 0)}
  child {node (travel) {Travel \\ (depth = 1)}
    child {node (transport) {Transport \\ (depth = 2)}
      child {node[fill=blue!15] (flight) {BookFlight \\ (depth = 3)}}
      child {node (train) {BookTrain \\ (depth = 3)}}
    }
    child {node (accom) {Accommodation \\ (depth = 2)}
      child {node[fill=red!15] (resort) {BookResort \\ (depth = 3)}}
      child {node (hotel) {BookHotel \\ (depth = 3)}}
    }
  }
  child {node (food) {Food \\ (depth = 1)}
    child {node (searchmenu) {SearchMenu \\ (depth = 2)}}
    child {node (orderfood) {Order \\ (depth = 2)}
      child {node (pizza) {OrderPizza \\ (depth = 3)}}
      child {node (sushi) {OrderSushi \\ (depth = 3)}}
    }
  };
\end{scope}
\node[draw, align=left, font=\footnotesize, anchor=north west,
      text width=0.3\linewidth, fill=gray!10, rounded corners,
      inner sep=6pt] at (-9, 2.5) {
\textbf{SIS Example Computation:} \\
\textbullet\ \textbf{Prediction:} \texttt{BookResort} \\
\textbullet\ \textbf{Ground Truth:} \texttt{BookFlight} \\
\textbullet\ \textbf{LCA:} \texttt{Travel} (depth = 1) \\
\textbullet\ $\mathrm{depth}(\texttt{BookResort}) = 3$ \\
\textbullet\ $\mathrm{depth}(\texttt{BookFlight}) = 3$ \\
\vspace{1mm}
$\Rightarrow \quad \mathrm{SIS} = \frac{2 \cdot 1}{3 + 3} = \frac{2}{6} = 0.33$
};
\draw[dashed, thick, gray] (flight) .. controls +(1.5,0.8) and +(-1.5,0.8) .. (resort);
\end{tikzpicture}}
\caption{Semantic Intent Similarity (SIS) example. Predicting \texttt{BookResort} instead of \texttt{BookFlight} yields a similarity score of 0.33 due to their shared ancestor \texttt{Travel}.}
\label{fig:sis-final}
\end{figure*}

\paragraph{Latency accounting.}
We report the symbolic layer's added latency separately from neural inference. Exact retrieval is $O(|V|m)$ for embedding dimension $m$; an approximate nearest-neighbor index makes lookup sublinear in practice. Prompt overhead is $O(|G_q|)$ labels and the decoding prior is $O(|T(G_q)|)$ token updates, both bounded by a small candidate set. This accounting keeps the deployment claim explicit without changing the underlying neural model.

\section{Experiments}
\paragraph{Setup.}
We evaluate on MultiWOZ 2.3 turns~\citep{budzianowski2018multiwoz,Han2021MultiWOZ}. The dataset is mapped to mobile-intent labels using an LLM annotator followed by ontology consistency checks. This setup is intended to test efficient multi-intent grounding, not to replace the standard MultiWOZ 2.3 dialogue-state leaderboard.

We compare TinyLlama \citep{zhang2024tinyllamaopensourcesmalllanguage} and Llama-3.2-3B \citep{meta2024llama32} with and without NOEM$^{3}$A. We report exact match (EM), micro Slot-F1 and the auxiliary hierarchy similarity described above. The most relevant comparison is within each compact model family, because those pairs isolate the cost and benefit of adding the symbolic layer. We set $k=5$ retrieved candidates and tune the decoding prior on development data, fixing $\beta=0.3$ and $\gamma=0.2$ after a small grid search found a broad plateau around those values.

\begin{table}[t]
\centering
\small
\setlength{\tabcolsep}{3.0pt}
\begin{tabular}{lccc}
\toprule
\textbf{Model / configuration} & \textbf{EM} & \textbf{Slot-F1} & \textbf{SIS} \\
\midrule
TinyLlama & 14.6 & 36.7 & 0.37 \\
TinyLlama + NOEM$^{3}$A & 23.2 & 55.3 & 0.62 \\
Llama-3.2-3B & 22.8 & 58.8 & 0.63 \\
Llama-3.2-3B + NOEM$^{3}$A & \textbf{35.7} & \textbf{72.8} & 0.85 \\
\midrule
Llama + SI only & 30.1 & 66.4 & 0.79 \\
Llama + LB only & 26.3 & 61.2 & 0.71 \\
Llama + SI + LB & 34.5 & 71.8 & 0.83 \\
Llama + SI + LB + CLF & 35.7 & 72.8 & 0.85 \\
\bottomrule
\end{tabular}
\caption{Compact-model results and Llama-3.2-3B ablations. SI is ontology retrieval plus prompting, LB is logit biasing and CLF is the auxiliary classifier. The last column is reported only as a hierarchy-aware diagnostic.}
\label{tab:main}
\end{table}

\paragraph{Accuracy gains under compact inference.}
Table~\ref{tab:main} shows that the ontology layer improves both compact models. TinyLlama gains 8.6 EM points and 18.6 Slot-F1 points. Llama-3.2-3B gains 12.9 EM points and 14.0 Slot-F1 points. The ablation indicates that retrieval plus ontology-conditioned prompting is the largest single contributor; logit biasing improves label fidelity, and the classifier adds a small final gain. This pattern is consistent with the intended efficiency role of the method: the ontology narrows the output space so that a compact model need not infer the full action inventory from parameters alone.

\paragraph{Latency overhead.}
Table~\ref{tab:latency} reports the added latency of the symbolic layer. Querying a 5k-node ontology took $2.1\pm0.3$ ms. Scaling the graph to 50k nodes increased lookup to 3.8 ms. The logit update touched at most 32 label tokens and remained below 0.1 ms. These numbers should not be read as universal device benchmarks, but they show that lookup and logit control add only millisecond-scale delay.

\begin{table}[t]
\centering
\footnotesize
\setlength{\tabcolsep}{4.5pt}
\begin{tabular}{lll}
\toprule
\textbf{Operation} & \textbf{Latency} & \textbf{Notes} \\
\midrule
5k lookup & $2.1\pm0.3$ ms & ANN/top-$k$ \\
50k lookup & 3.8 ms & scaled index \\
Logit mask & $<0.1$ ms & $\leq$32 tokens \\
\bottomrule
\end{tabular}
\caption{Measured symbolic-layer latency. Retrieval and logit control are separated from neural decoding.}
\label{tab:latency}
\end{table}

\paragraph{Prompt-template check.}
We also tested whether gains depend on the exact wording of the candidate-list prompt. Five development-set variants changed label order, added a task cue, inserted one example, replaced \emph{intent} with \emph{option} or removed most meta-language. EM stayed within roughly one point of the baseline prompt, suggesting that the main signal is the retrieved label set rather than a fragile phrasing choice.

\section{Discussion}
\paragraph{Why this helps efficiency.}
The method does not make the language model itself cheaper. Instead, it improves the amount of task accuracy obtained from a small model by providing a compact symbolic prior. This is useful for mobile agents because supported actions are usually known in advance: the ontology can double as the planner's schema and the NLU model's constrained label space. In other words, the system spends a small amount of compute to recover the valid action vocabulary before asking the model to decide among actions. This is a better match to mobile agents than open-ended generation over arbitrary text.

\paragraph{Failure localization.}
The symbolic interface makes errors easier to inspect. If an intent is missed, developers can check whether it was absent from retrieval, present but ignored during generation, or filtered by the classifier. These failure modes suggest different fixes: improving node descriptions, adjusting retrieval expansion, tuning $\beta,\gamma$ or changing merge thresholds. This is also why we keep the hierarchy-aware score as a diagnostic: it can reveal whether the system selected the wrong leaf inside a relevant branch or moved to an unrelated branch, but it should not replace EM or Slot-F1.

\paragraph{Relation to efficient NLP.}
The approach is complementary to compression, pruning and mobile-specific models~\citep{Sun2020,Yi2024PhoneLM,Tan2024}. Those methods reduce neural cost directly; ontology guidance reduces the need to scale model size for a structured intent task. The two strategies can be combined: a quantized or mobile-specialized model can still benefit from ontology retrieval, while the ontology layer remains independent of the model's internal architecture.

\section{Conclusion}
We presented NOEM$^{3}$A{} as an efficient ontology-guided layer for multi-intent understanding in mobile agents. The central contribution is practical rather than metric-centric: a small symbolic lookup and ontology-guided logit biasing can improve compact local models while adding only millisecond-scale latency. The formalization above makes the mechanism inspectable, and the ablation indicates where gains come from. Future work should evaluate the same mechanism on real mobile-agent traces, richer GUI context and broader hardware profiles.

\section*{Limitations}
Our evaluation uses the MultiWOZ 2.3 dataset rather than a naturally collected mobile-agent corpus. The annotations and ontology require human auditing; poor ontology design can affect retrieval and the auxiliary hierarchy diagnostic. Logit biasing assumes that intent labels map to short token spans, so long labels may require special tokens or classifier-only handling. The latency observations come from one device class and should be replicated across hardware and runtime stacks before deployment claims are generalized.

\section*{Ethical considerations}
The system is designed for privacy-preserving local intent understanding, but mobile agents can execute consequential actions. Deployments should require confirmation before purchases, bookings, data sharing or other high-impact actions. Ontology construction can encode omissions or cultural assumptions, so the intent inventory and LLM-assisted annotations should be audited before production use.

\setlength{\bibsep}{0pt plus 0.2ex}
{\footnotesize\bibliography{custom}}

\end{document}